\documentclass[11pt]{article}

\usepackage[]{ACL2023}

\usepackage{times}
\usepackage{latexsym}
\usepackage{booktabs}
\usepackage{graphicx}
\usepackage{caption}
\usepackage{subcaption}
\usepackage{amsmath}
\usepackage{amssymb}
\usepackage{multirow}
\usepackage{cleveref}

\usepackage[T1]{fontenc}

\usepackage[utf8]{inputenc}

\usepackage{microtype}

\usepackage{inconsolata}

\title{Detection and Measurement of Syntactic Templates in Generated Text}

\author{\textbf{Chantal Shaib$^1$}\quad
\textbf{Yanai Elazar$^{2,3}$}\quad 
\textbf{Junyi Jessy Li$^4$}\quad
\textbf{Byron C. Wallace$^1$}\quad\\
$^1$Northeastern University, $^2$Allen Institute for AI, \\$^3$University of Washington, $^4$The University of Texas at Austin  \\
\small\texttt{\{shaib.c, b.wallace\}@northeastern.edu}\\
}

\begin{document}
\maketitle

\begin{abstract}
The diversity of text can be measured beyond word-level features, however existing diversity evaluation focuses primarily on word-level features. 
Here we propose a method for evaluating diversity over syntactic features to characterize general repetition in models, beyond frequent $n$-grams. Specifically, we define \textit{syntactic templates} (e.g., strings comprising parts-of-speech) and show that models tend to produce templated text in downstream tasks at a higher rate than what is found in human-reference texts.
We find that most (76\%) templates in model-generated text can be found in pre-training data (compared to only 35\% of human-authored text), and are not overwritten during fine-tuning or alignment processes such as RLHF.  
The connection between templates in generated text and the pre-training data 
allows us to analyze syntactic templates in models where we do not have the pre-training data.
We also find that templates as features are able to differentiate between models, tasks, and domains, and are useful for qualitatively evaluating common model constructions.
Finally, we demonstrate the use of templates as a useful tool for analyzing style memorization of training data in LLMs \footnote{\url{https://cshaib.github.io/syntactic_templates/}}.
\end{abstract}

\section{Introduction}
An open question about large language models (LLMs)
is what patterns such models learn from pre-training data \citep{goldberg2019assessing,petroni2019language,bender2021dangers,chen2024skill}, and whether the same patterns appear generally across downstream tasks and datasets \citep{Hupkes2023}. 
While prior work has focused on the quality of generation \citep{zhang2019bertscore,dou2022gpt,kryscinski2020evaluating}, and more recently
on text generation novelty \citep{mccoy2023much,ngram-novelty}, there has been limited work on characterizing the sorts of lexical patterns that are learned by LLMs. 

\begin{figure}
    \centering
    \resizebox{0.48\textwidth}{!}{
    \includegraphics{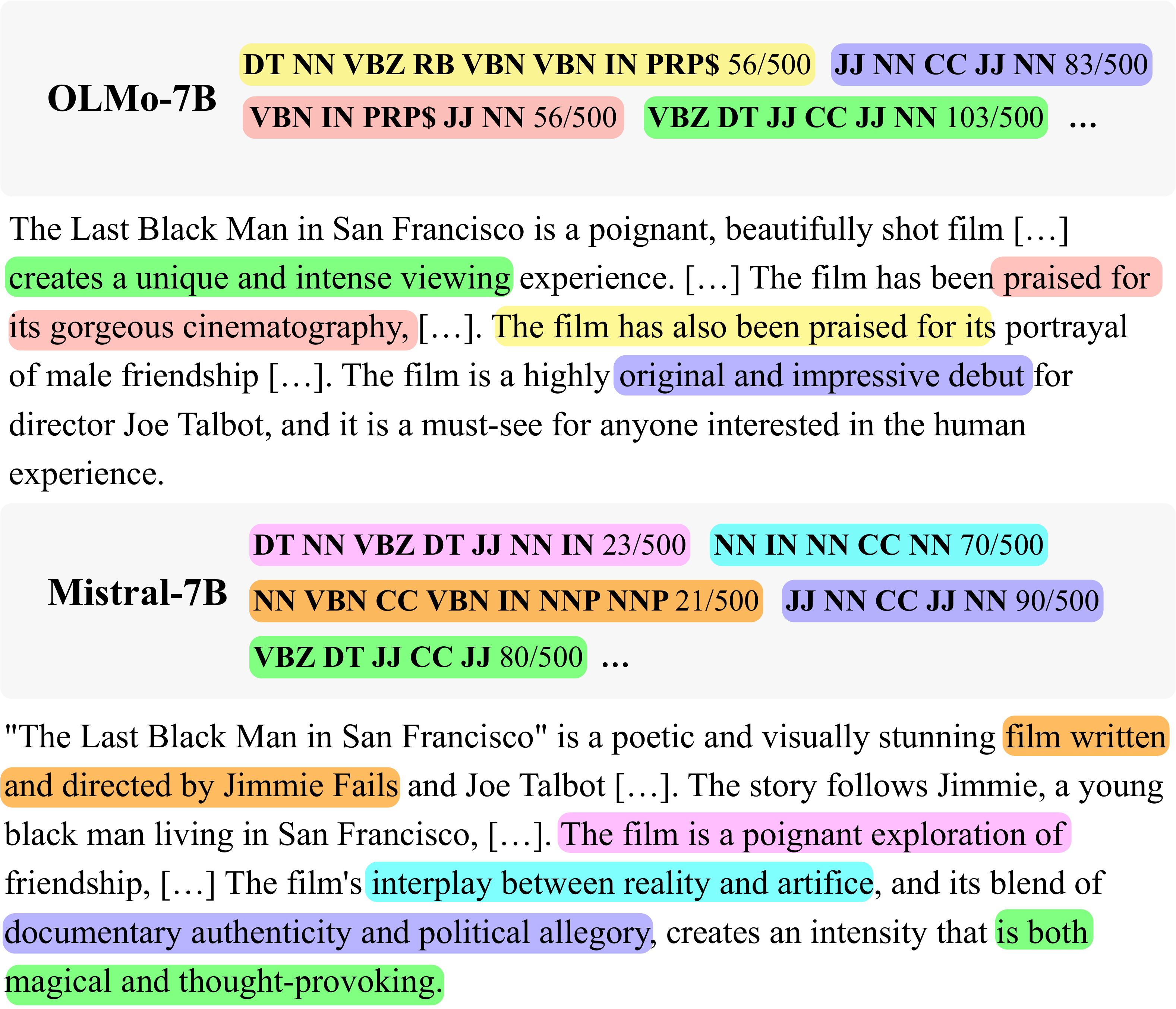}
    }
    \caption{Sample movie meta-reviews generated by OLMo-7B (top) and Mistral-7B (bottom) by prompting the Rotten Tomatoes dataset. Templates appear at varying rates (frequency shown out of 500 generations), and differ across models. We extract templates from the entire corpus of generated text for each model, and match the text to the part-of-speech templates (highlighted), following by the frequency of each template. 
    }
    \label{fig:templates}
\end{figure}

Consider, for instance, the generated text from OLMo-Instruct in \Cref{fig:templates}, which is sampled from a corpus of movie review summaries. 
This was produced by prompting the model to summarize a collection of human-written movie reviews: ``\emph{The Last Black Man in San Francisco is a poignant, beautifully shot film [...] creates a unique and intense viewing experience [...]}''. 
While this generated text was not seen in Dolma \cite{soldaini2024dolma}, OLMo's pre-training data,
we find a total of 35 unique repeated sequences of part-of-speech (POS) tags of lengths $n = 5$ to $8$ in the summarized movie reviews. 
Further, we find that 33 out of the 35 (95\%) sequences appear in the pre-training data.
As such, while the generated text itself is novel, it relies on common syntactic sequences seen
in the training data.

In this work, we quantify and measure LLMs' usages of repetitive sequences in text generation. We introduce and focus on \emph{syntactic templates}, namely POS sequences, a syntactic abstraction over texts. 
We seek to answer the following questions:

\vspace{0.7em}
\noindent {\bf RQ1} To what extent do outputs generated by instruction-tuned LLMs contain templates? 

\vspace{0.4em}
\noindent {\bf RQ2} Can we locate model-generated templates in (pre-)training data?

\vspace{0.4em}
\noindent {\bf RQ3} Can syntactic templates be used for detection of data memorization? 
\vspace{0.7em}

We start by introducing syntactic templates, and defining methods for detecting and measuring such templates in generated texts (\S \ref{sec:detecting}).
We evaluate eight models on three different tasks (\S \ref{sec:templates-in-text}).  
We show how training data templates are memorized and subsequently generated by models trained on them (\S \ref{sec:templates-training-data}). We then show how such insights allow one to draw conclusions about the training data used by closed models in a downstream summarization task (\S  \ref{sec:templates-in-text-closedsrc}).
Finally, we show that our metrics can also be used as a softer version of memorization. For instance, while \citet{carlini2022quantifying} estimates that 1\% of texts to be memorized,
we find between 0.8 - 3.1\% more \textit{soft-memorized} texts over verbatim memorization,
often by replacing numbers and synonyms (\S\ref{sec:memorization}).

\section{Related Work}

\paragraph{Diversity in Text Generation} 
Past evaluations of diversity in LLM outputs have primarily focused on token-level diversity \citep{Montahaei2019JointlyMD, Bache2013TextbasedMO}. Diverse sampling strategies have been introduced to address the lower token diversity observed in neural text generation \citep{Holtzman2020The, roberts2020decoding}, however it is unclear whether such sampling strategies also increase the diversity of the syntactic structure in LLMs.
Beyond lexical diversity, \citet{padmakumar2023does} extend definitions for measuring content diversity, which has broad applications in downstream tasks such as summarization and creative story generation. 
Recent work has quantified the drop in generated-text diversity specifically relative to the RLHF training process, however this again focused primarily on token level diversity \citep{kirk2024understanding}.
Our work aligns more closely with the first body of work; we measure the syntactic structure of text rather than its semantic content. Most similar to our work is \citet{bar2012text}, which broadly evaluates text repetition metrics at the stylistic, content, and lexical level. Our methods do not address repetition in content but rather focus on extending the characterization of lexical and stylistic repetition with text abstractions in LLMs.

\paragraph{Structural Analysis of Text}
\citet{dimarco-hirst-1993-computational} provide a computational approach comprising lexical and syntactic components to describe stylistic elements in model-generated text. The discussion around style in writing has been adopted broadly for a variety of downstream tasks such as author or model attribution \citep{wu2023survey, Lample2018MultipleAttributeTR, rosenfeld2024whose}. While our main goal is to provide measurements and characterizations of repetitive syntactic features in text,  definitions of stylistic elements are closely related and help contextualize our findings. One can use our definitions of templates to ask broader questions about the prevailing syntactic style in a given corpus. Indeed, recent works adopt various measures of linguistic analysis to address differences in writing style in both human-written and model-generated texts \citep{krishna2020reformulating, soler-company-wanner-2017-relevance}.

\paragraph{AI Text Detection}
In identifying $n$-gram features that appear in high frequencies in model-generated text, a natural question arises as to whether such features can be used to reliably \emph{detect} model-generated text. Prior work has established that this is difficult, and that text-level features at the corpus level correlate with text being model-generated \citep{liang2024monitoring, liang2024mapping}. In this work, we make no claims for the use of templates in AI-text detection. Our aim is to \emph{characterize} patterns rather than \emph{detect} generated outputs, and to provide a basis for future work on model linguistic diversity.    

\section{Detecting Syntactic Templates}
\label{sec:detecting}
Our goal is to search for abstract representations of texts to capture more subtle repetitions than mere text memorization. 
Repeated strings of literal tokens may not be sufficient for describing such redundancy nor why a summary produced by, e.g., ChatGPT, might seem familiar. 

Focusing on syntactic patterns rather than tokens allows us to capture such repetitions
For example, a pattern consisting of the part-of-speech sequence \texttt{DT JJ NN IN DT JJ NN} will match to phrases in movie reviews (\textit{``a romantic comedy about a corporate executive''}) and in news summarization (\textit{``a humorous insight into the perceived class''}), even though these sentences have only one token overlap.

\subsection{Defining Templates}
\label{sec:definition}
Given a sequence of tokens \( T = (t_1, t_2, \ldots, t_n) \), and a function $f$ that computes an abstraction over $T$ (e.g.,  part-of-speech tags), we define a \emph{template} as a sub-sequence of abstractions over the tokens \( f(T) \) that repeats at least $\tau$ times in $T$. 
\Cref{fig:templates} shows examples of templates and their counts across the Rotten Tomatoes dataset \citep{leone2020rotten}.

\subsection{Extracting Syntactic Templates from Text}
\label{method:extract_temps}

We operationalize the definition in~\ref{sec:definition} as parts-of-speech (POS). 
For sub-sequences of POS we consider templates of length $n \in \{4,5,6,7,8\}$.
Templates are characterized by their high frequency across the texts in a given corpus (e.g., one comprising texts generated by a particular LLM) 
Practically, we choose $\tau$ relative to the sample size. 
For Rotten Tomatoes we retain the top 100 most common template where the least frequent template appears 4 times in the dataset. 

\begin{figure}
    \centering
    \includegraphics[width=0.9\linewidth]{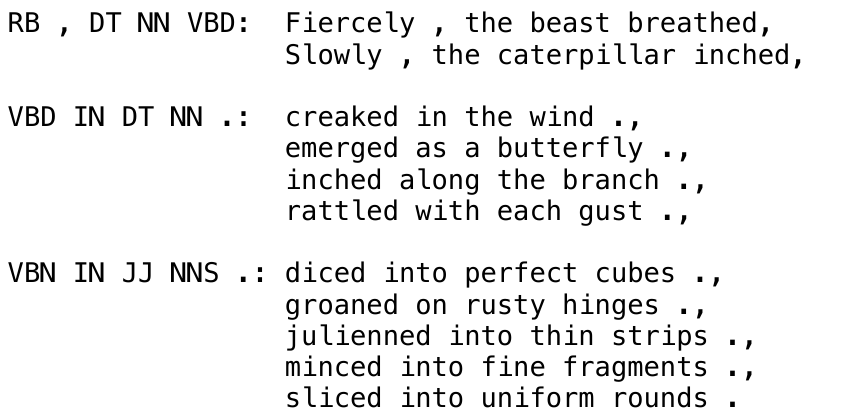}
    \caption{Example templates (left) and matched text (right) returned from the \texttt{diversity} package.}
    \label{fig:ex_method}
\end{figure}

To extract templates we use \texttt{diversity} \citep{shaib2024standardizing},\footnote{\url{https://pypi.org/project/diversity/}} a library providing tools to evaluate token and POS diversity in a dataset. We use this tool to first tag all tokens in a corpus with their corresponding POS tags, then search for the top 100 most frequent $n$-grams across these tags. \texttt{diversity} uses the SpaCy POS tagger,\footnote{\url{https://spacy.io/api/tagger}} which relies on the Penn Treebank set of 36 tags \citep{taylor2003penn}. 
After tagging, we return frequent $n$-grams of POS, the corresponding matched text. Figure~\ref{fig:ex_method} illustrates the output of running the template extraction process. 

The pipeline for identifying templates can be further extended to other tagging libraries and types. For example, we also explore constituency parses as an alternative to POS tags. 
Extracting templates and matching over tokens is non-trivial for constituency parses. We provide examples of the patterns identified by this abstraction in \Cref{appx:consituents} and leave further analysis to future work.

\subsection{Metrics for Measuring Templates}
Our goal for extracting templates is to assess and characterize different levels of repetition in LLMs. 
We calculate three metrics using templates, 
(1) the diversity of the POS tags that are generated using CR-POS (2) the fraction of texts generated with a template using template rate and count, and (3) the number of templates that appear within each text using templates-per-token.
We now elaborate on each one.

\paragraph{CR-POS.} 
At the most granular level, we are interested in quantifying the $n$-gram diversity of the POS tag sequences present in the text. 
Lossless text compression algorithms---such as gZip---are optimized to detect repeated characters in sequences, and rely on this to compress documents without any loss of information. 
If a document contains frequent repeated strings, the document will be more compressible, resulting in a larger difference in compressed size relative to the original document size. \citet{shaib2024standardizing}
show that using gZip to calculate a compression ratio (CR) can provide an efficient measure for capturing lexical diversity, specifically $n$-gram repetition. 

We calculate CR over a set of POS-tagged text, with higher values indicating that text is highly compressable (and therefore shows lower diversity). To calculate the CR, we concatenate all POS-tagged text into a sequence, and measure the ratio between the original document size and the compressed document size: 

\begin{equation}
    \label{eq:cr-pos}
    \text{CR}(f(T\oplus)) = \frac{ |f(T\oplus)|}{\text{compressed } (|f(T\oplus)|)}
\end{equation}

\noindent Where $T\oplus$ is the concatenated sequence of text, and $f(T\oplus)$ is POS-tagged text. 
Higher compression ratios imply more redundancy in the text, and therefore lower diversity of the sequence. 

\paragraph{Template Rate}
We measure the fraction of texts in a corpus (sequence) that contain at least 1 template to quantify how frequently templates appear across an entire corpus. 
\begin{equation}
    \label{eq:template_indicator}
    TR = \frac{1}{T} \sum_{i=1}^{T} I_i
\end{equation}

Where, $T$ is the sequence of text (corpus), and
\begin{align*}
    I_i &= \begin{cases}
        1 & \text{if text } i \text{ contains at least 1 template} \\
        0 & \text{otherwise}
    \end{cases}
\end{align*}

\paragraph{Templates-per-Token} 
In practice, text can contain many templates. Measures of  diversity are confounded by text length \citep{salkar-etal-2022-self}, which also applies to template counts; if a model tends to produce longer texts, there is a higher chance that any given output will contain a template. 
To compare between text sources,
we can length normalize: 
\vspace{-0.15em}
\begin{equation}
    \label{eq:templates-per-token}
    \text{TPT}(T\oplus) = \frac{\frac{1}{T}\sum_{i=1}^{T}\text{\# templates in $t_i$}}{\frac{1}{T}\sum_{i=1}^{T}\text{\# words in $t_i$}}
\end{equation}

Where $T$ is a concatenated sequence of tokens forming a corpus for a particular text source, and $t$ the string. 

\section{Experimental Setup}

\subsection{Models}
\paragraph{Open Models}
We first evaluate the incidence of templated text in two open-ended generation tasks using OLMo-7B Instruct \citep{Groeneveld2024OLMoAT}, a fully open source model that released model training checkpoints and its training data. This allows us to evaluate templates in its 
training datasets: Dolma \citep{soldaini2024dolma}, Tulu-V2 \citep{Ivison2023CamelsIA}, and Ultra-feedback \citep{cui2023ultrafeedback}.

We then evaluate templates across closed source models (which do not release training data), specifically: Mistral, Llama (-2, -3), Alpaca, and GPT-4o.  

\paragraph{Fine-tuned (Instruction) Models} 
We experiment with a total of 8 instruction-tuned models. We use Mistral (Instruct, 7B; \citealt{Jiang2023Mistral7}), Alpaca (7B, 13B; \citealt{alpaca, Wang2022SelfInstructAL, Touvron2023LLaMAOA}). 
In addition, 5 models are further trained on human preferences: 
OLMo (Instruct, 7B; \citealt{Groeneveld2024OLMoAT}),
Llama-2 (Chat-HF, 7B-70B; \citealt{Touvron2023Llama2O}), and Llama-3 (Instruct, 70B \citealt{dubey2024llama3herdmodels}).

\subsection{Decoding} While greedy decoding is a common decoding strategy for many popular downstream generation tasks, one can explicitly control token diversity at inference time via choice of decoding hyperparameters such as temperature. We evaluate generation under various decoding strategies and model sizes.
We refer to \citet{Wiher2022OnDS} for an in-depth discussion on the impact of sampling on generated text, and here focus specifically on varying hyperparameters and resultant impact of the appearance and frequency of templates. For the former, we use greedy decoding, and separately vary temperature and top-$p$ for decoding with sampling. Top-$p$ (nucleus) sample restricts the subset of tokens such that the combined probability reaches a threshold $p$ \citep{Holtzman2020The}.

\subsection{Tasks and Datasets}
\paragraph{Open-Ended Generation} To evaluate intrinsic template behaviour we evaluate open-ended generation tasks in two settings. 
In the first setting, we sample generations from the model given only a special token denoting beginning of sequence (\texttt{[BOS]}). 
In the second, we randomly sample 100 tokens from Dolma and use these tokens to prompt further open-ended generation from the model.

\paragraph{Synthetic Data Generation} LLMs are increasingly used to create synthetic training datasets, which are often used to train downstream models  (e.g., \citealt{Wang2022SelfInstructAL}). 
We evaluate templates in Cosmopedia, a synthetic dataset generated by prompting  Mixtral-8x7B-Instruct with instructions to produce text relating to textbooks, blogposts, stories, posts and WikiHow articles \citep{benallal2024cosmopedia}.
We prompt OLMo-7B with the Cosmopedia instructions and evaluate the resulting generations. 

\begin{table}
\begin{centering}
\resizebox{0.5\textwidth}{!}{

\begin{tabular}{llr|lr}
  &  \multicolumn{2}{c}{\textbf{Open Generation}}&  \multicolumn{2}{c}{\textbf{Rotten Tomatoes}}\\
\toprule
 \textbf{\begin{tabular}[c]{@{}l@{}} Decoding\\Strategy\end{tabular}}& \textbf{\begin{tabular}[c]{@{}l@{}} CR: \\POS\end{tabular}}&\textbf{\textbf{\begin{tabular}[c]{@{}l@{}}$\geq1$ Templates\\\% ($n=6$)\end{tabular}}}    &  \textbf{\begin{tabular}[c]{@{}l@{}} CR: \\POS\end{tabular}}& \textbf{\textbf{\begin{tabular}[c]{@{}l@{}}$\geq1$ Templates\\\% ($n=6$)\end{tabular}}}    \\\midrule
 Greedy& 702.8& 100.0 (0.065)& 6.45&97.0 (0.041) \\ 
 Default & 5.81&75.5 (0.009)&  6.33& 96.6 (0.041)\\\midrule
\texttt{temp 0.8} & 6.74&71.8 (0.007)&  6.26& 96.6 (0.043)\\
\texttt{temp 0.85}& 6.48& 74.0 (0.007)& 6.22&96.6 (0.041)\\
\texttt{temp 0.9}& 6.22& 73.4 (0.009)& 6.17&97.4 (0.039)\\
\texttt{temp 0.95}& 5.98& 75.4 (0.010)& 6.14&97.2 (0.040)\\\midrule
\texttt{top\_p 0.8}& 7.03& 76.5 (0.007)& 6.31&97.8 (0.041)\\
\texttt{top\_p 0.85}& 6.71&71.0 (0.007)&  6.27& 96.6 (0.041)\\
\texttt{top\_p 0.9}& 6.50&75.3 (0.008)&  6.22& 96.2 (0.039)\\
\texttt{top\_p 0.95}& 6.17&77.2 (0.009)&  6.31& 95.8 (0.041)\\\bottomrule
\end{tabular}
}
\caption{Compression ratio with POS (CR-POS), average text length and percentage of generated outputs with at least 1 template of size $n=6$, when varying \texttt{temperature} and \texttt{top\_p} for OLMo-7B decoding. Arrows indicate higher template rates. 
}
\label{tab:cr_pos_sample}
\end{centering}

\end{table}

\begin{table}[t]
\begin{centering}
\resizebox{0.3\textwidth}{!}{
\begin{tabular}{lll}
\toprule
 \textbf{Dataset}& \textbf{\begin{tabular}[c]{@{}l@{}} CR: \\POS\end{tabular}}&\textbf{\textbf{\begin{tabular}[c]{@{}l@{}}$\geq1$ Templates\\\% ($n=6$)\end{tabular}}}   \\\midrule
  
 Dolma& 5.65&82.6 (0.012)\\ 
  Cosmopedia& 5.76 &99.1 (0.014)\\\bottomrule
\end{tabular}
}
\caption{CR-POS, template-per-token, and template counts for templates of size $n=6$ reported for OLMo-7B text generated with Cosmopedia Instructions, and 100 sampled tokens from the Dolma dataset, with greedy decoding.}
\label{tab:cr_pos_cosmopedia}
\end{centering}

\end{table}

\paragraph{Summarization} 
Summarization is a common benchmark for long text generation. 
We evaluate models on a handful of summarization datasets, including single- and multi-document tasks.
Such datasets allow us to study templates in longer sequences that would not be evident in tasks where only a few tokens are generated. 
For general English-language tasks, we generate summaries and reviews over news (CNN/Daily Mail; \citealt{nallapati2016abstractive}), movies (Rotten Tomatoes; \citealt{leone2020rotten}), and books (BooookScore; \citealt{Chang2023BooookScoreAS}). 

We also look at templates in the biomedical domain as an example of a domain-specific task. 
Cochrane is a dataset of systematic reviews
summarizing the evidence over
medical interventions \citep{Wallace2020GeneratingN}. We prompt models to generate systematic reviews.
Importantly, these datasets include human-written reference summaries, which serve as a baseline to compare our task-specific template analysis.

\section{Templates in Model-Generated Text}
\label{sec:templates-in-text}
We first evaluate OLMo-7B Instruct on 3 tasks: open-ended generation, synthetic data generation, and summarization, using both greedy and varying temperature and top-$p$
sampling strategies (\textbf{RQ1}). 

Table~\ref{tab:cr_pos_sample} shows the effect of varying sampling hyperparameters temperature and top-$p$ on the overall diversity of the generated text with OLMo-7B with open-generation and summarization.

Varying sampling strategies in the open-generation task results in a higher variance of template rates (74.4\% $\pm$ 2.1) compared to templates rates in the summarization task (96.8\% $\pm$ 0.6). These results indicate that templatic text in summarization appears in spite of sampling strategies intended to increase (lexical) diversity.
Overall, the rate of templates is much higher in the Rotten Tomatoes dataset than for open-generation, indicating downstream tasks such as summarization, which often entail prompting with instructions, may yield more repetitive structures.\footnote{Note that we show the incidence of templates given different instructions in Appendix~\ref{appx:prompts}}

Table ~\ref{tab:cr_pos_cosmopedia} shows the rate of templates on two additional tasks: Synthetic data generation and data generation with Dolma.
Cosmopedia results in a higher incidence of templates (99.1\%) and templates per token (0.014), compared to Dolma (82.6\%, 0.012).

\section{Searching For Templates in Pre-training Data}
\label{sec:templates-training-data}
One hypothesis for the emergence of templates in generated text is that these templates are over-represented in the training data \textbf{(RQ2)}.
Here we interrogate this empirically. 

\subsection{Emergence of Templates in Training}
\label{sec:modeling-templates}
We first aim to understand when during training models start to generate templates.
We measure the perplexity of matched texts from a set of previously extracted templates (following \S \ref{method:extract_temps}) across OLMo's checkpoints. 
Higher perplexity values indicate the templates are assigned low likelihood at that checkpoint.

For each model checkpoint, we average the perplexities of templates of length $n=6$ and compare to the perplexities of randomly sampled 6-grams. 
We calculate the average perplexity for the dataset using:  
\vspace{-0.5em}
\begin{equation}
    \frac{1}{|D|}\frac{1}{|N|}\sum_{j=1}^{|D|}\sum_{k=1}^{|N|} 2 ^{H(p_k)}
\end{equation}

\noindent Where $N$ is the total number of templates in the document, and $D$ the total number of documents in the dataset
We repeat this process for randomly sampled 6-grams (distinct from the templates) to match the number of templates.

\begin{figure}
    \centering
    \resizebox{0.5\textwidth}{!}{
    \includegraphics{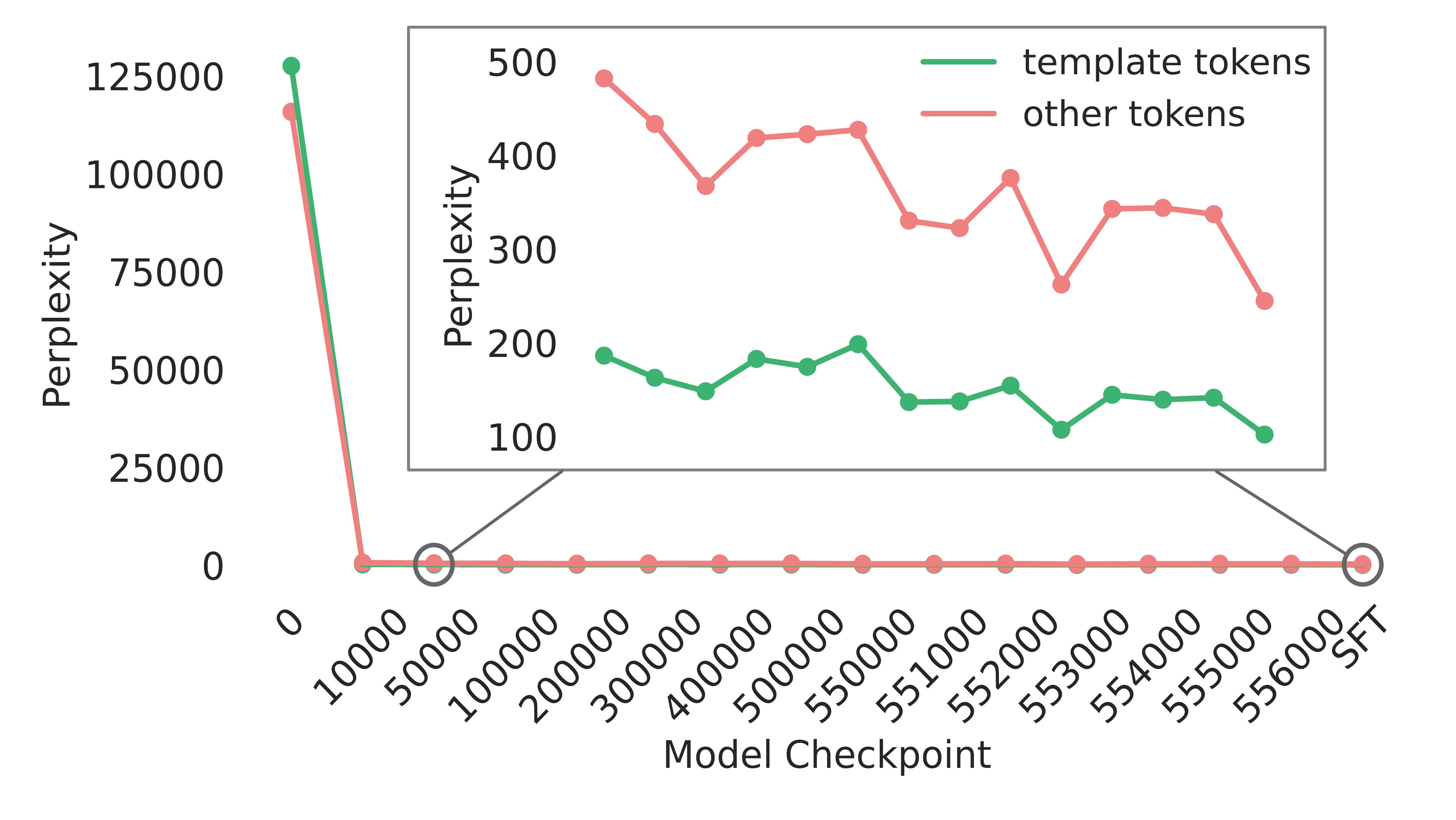}
    }
    \caption{Perplexity of matched template at different model checkpoints. Templates initially have higher perplexity than other tokens, but quickly drop after initial training steps.} 
    \label{fig:prps}
\end{figure}

\begin{figure}[t]
    \begin{minipage}{0.45\textwidth}
        \centering
        \includegraphics[width=\textwidth]{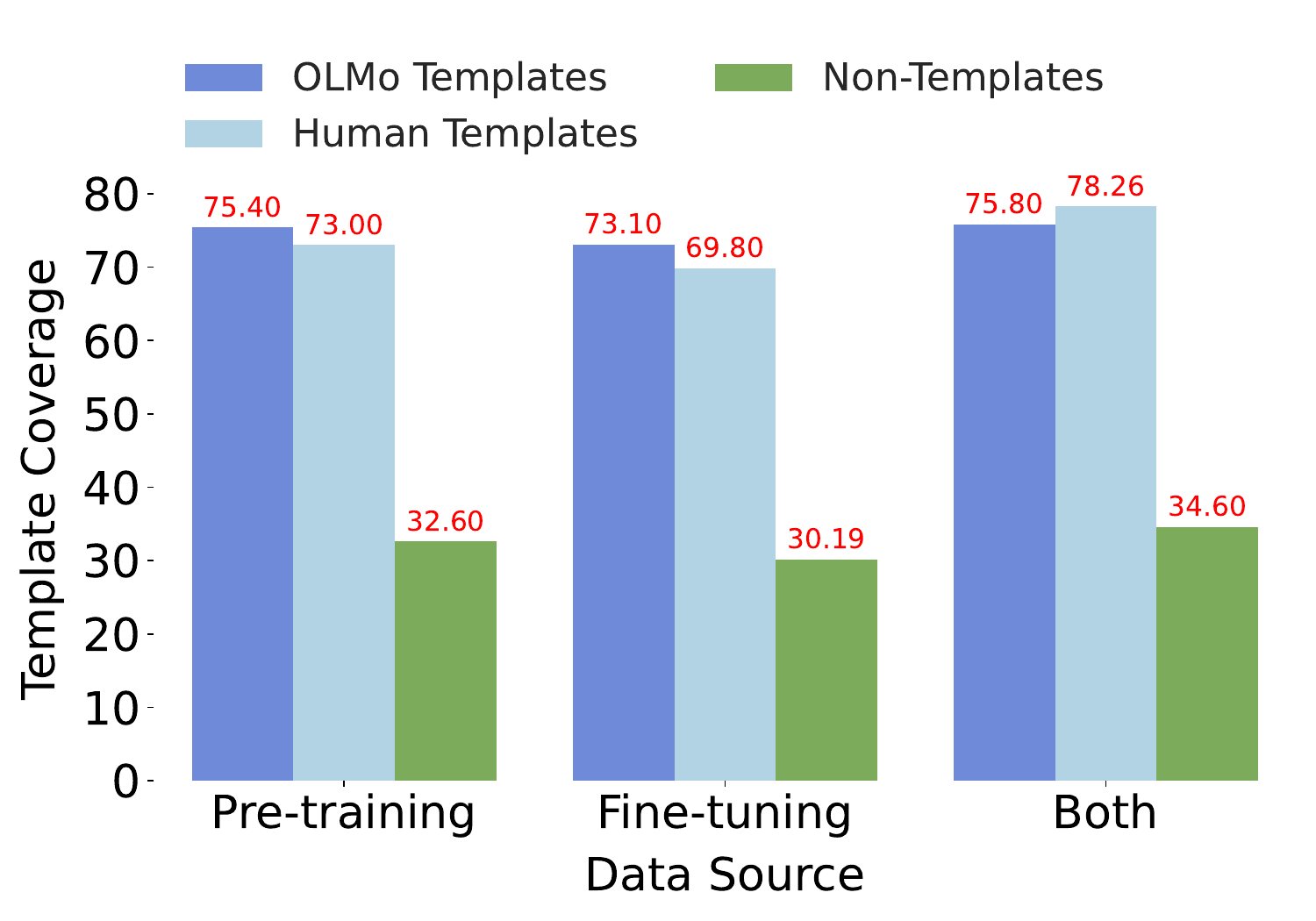}
        \caption{Coverage of templates found in pre-training data, fine-tuning data, and both datasets combined. Templates are found at a much higher rate in the training data than random n-gram sequences.}
        \label{fig:pretrain_coverage}
    \end{minipage}\hfill
    \begin{minipage}{0.45\textwidth}
        \centering
        \includegraphics[width=\textwidth]{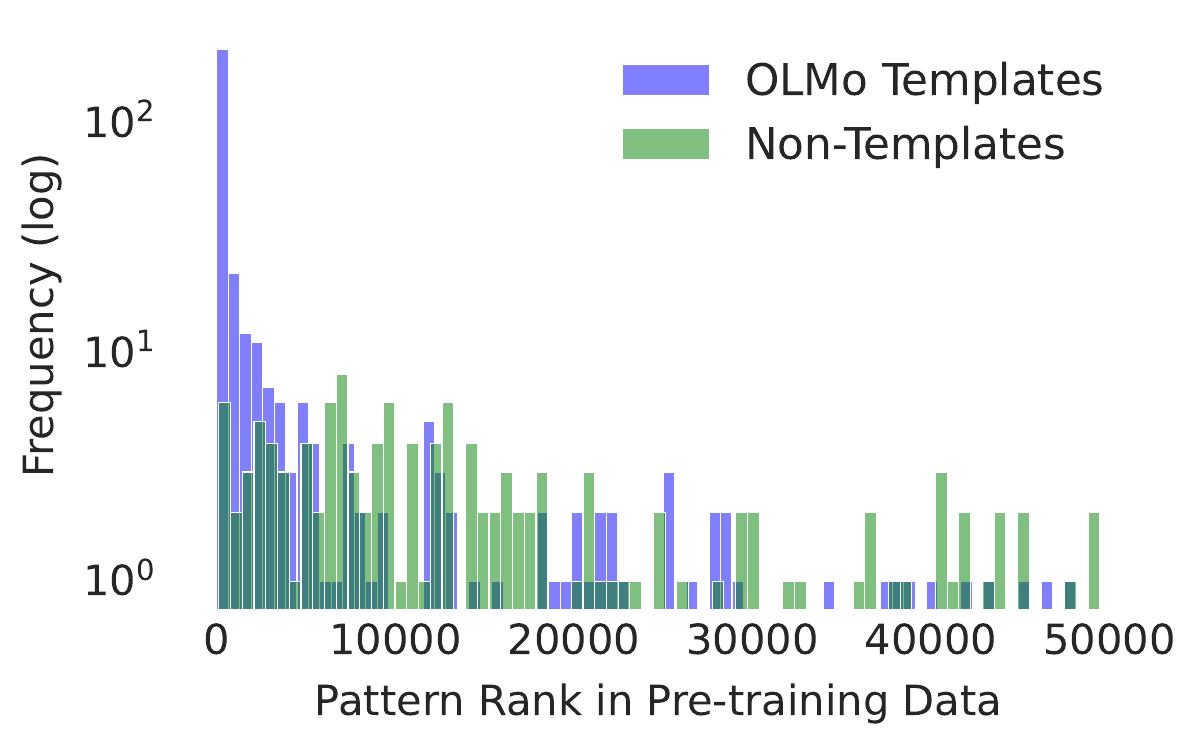}
        \caption{Ranking of OLMo templates and non-templated sequences in frequency of pre-training data. Templates appear at higher ranks (and are therefore more frequent) than non-templated sequences.}
        \label{fig:pretrain_rank}
    \end{minipage}
    
\end{figure}

\paragraph{Results} Figure~\ref{fig:prps} shows the average perplexities across model checkpoints. We find that templates are learned quickly---by the first model checkpoint (which was trained on 4B tokens). Average perplexity drops to around 500 for non-template tokens, compared to 200 for templates. 

These findings are surprising, and indicate that templates are learned early in pre-training, rather than during the fine-tuning process. The average perplexities remain lower for template tokens for the remainder of the training process.

\subsection{Templates in Pre-training Data}
\label{sec:pre-train-data}

The lower perplexities in the above finding indicates that templates are seen fairly early on in pre-training compared to non-templated sequences of training data text. 
We next measure the incidence and types of templates in the pre-training data, and whether they correspond to the templates that models produce.

To search for template coverage by OLMo, we start by selecting a random subset of the Dolma dataset, containing 
10 billion tokens.
We then annotate all of the sequences with a POS tagger using the Dolma toolkit \citep{soldaini2024dolma}.
Finally, we find the 50K most common POS-grams in the data for sequence length of six using the \textsc{WIMBD} toolkit \citep{elazar2023s}, which is optimized for search and count at large scales.

\paragraph{Results} Figure~\ref{fig:pretrain_coverage} 
 shows the coverage of templates produced by OLMo in the pre-training data, the fine-tuning data, and their concatenation. 
 We find that 75\% of templates produced by OLMo are found in the pre-training data, indicating that a majority of templates are not a novel construction learned during fine-tuning.
 Rather, they are learned directly from pre-training data. 
In comparison, only 34\% of randomly sampled non-templated sequences are found in the pre-training data.
Further, Figure~\ref{fig:pretrain_rank} shows that the templates OLMo generates consistently rank higher in frequency in the pre-training dataset, compared to randomly sampled non-templates. 
The difference in ranks between the templates and non-templates is statistically significant; the median rank in templates and non-templates are 337.5 and 9651.0 (Mann--Whitney $U = 6043$, $p < 0.05 \text{ two-tailed}$). Overall we find that, not only do most of the templates produced in downstream generation tasks appear in the pre-training data, but are also often very frequent sequences in the pre-training data. 

\begin{table*}[t]
\begin{centering}
\resizebox{0.75\textwidth}{!}{
\begin{tabular}{l@{}llr|lr|lr}
 & & \multicolumn{2}{c}{\textbf{\begin{tabular}[c]{@{}l@{}} Rotten Tomatoes\end{tabular}}}& \multicolumn{2}{c}{\textbf{Cochrane}}& \multicolumn{2}{c}{\textbf{CNN/DM}}\\
\toprule
 &\textbf{Model}& \textbf{\begin{tabular}[c]{@{}l@{}} CR: \\POS\end{tabular}}&\textbf{\textbf{\begin{tabular}[c]{@{}l@{}}$\geq1$ Templates\\\% ($n=6$)\end{tabular}}}   & \textbf{\begin{tabular}[c]{@{}l@{}} CR: \\POS\end{tabular}}& \textbf{\textbf{\begin{tabular}[c]{@{}l@{}}$\geq1$ Templates\\\% ($n=6$)\end{tabular}}}   & \textbf{\begin{tabular}[c]{@{}l@{}} CR: \\POS\end{tabular}}&\textbf{\textbf{\begin{tabular}[c]{@{}l@{}}$\geq1$ Templates\\\% ($n=6$)\end{tabular}}}   \\\midrule
  &
  Reference& 5.31&46.4 (0.040) & 5.63& 83.3 (0.049)& 5.33&36.0 (0.013)\\
  &Input Documents& 5.82&29.3 (0.001) & 5.96& 98.5 (0.021)& 5.54&98.4 (0.020)\\ \midrule
  \ \ 
  &
  OLMo-7B     & 6.45&97.0 (0.041) & 6.53& 74.0 (0.030)& 5.83&\textbf{91.2 (0.025)}\\
 &Mistral-7B  & 6.29&\textbf{99.6 (0.043)}& 6.10& 99.5 (0.043)& 5.70&\textbf{89.9 (0.029)}\\
 &Llama-2-7B  & 6.87&\textbf{93.0 (0.047)}& 6.43& 88.4 (0.042)& 5.71&\textbf{90.4 (0.028)}\\
 &Llama-2-13B &                                                                   6.70&\textbf{99.0 (0.060)}& 6.65& \textbf{95.1 (0.052)}& 5.91&\textbf{97.4 (0.042)}\\
 &Llama-2-70B &                                                                   6.36&\textbf{99.3 (0.123)}& 6.51& 99.7 (0.042)& 5.69&\textbf{87.4 (0.027)}\\
 &Llama-3-70B&     6.39&          \textbf{99.2 (0.151)}& 6.50& 99.5 (0.030)& 5.66&\textbf{83.2 (0.024)}\\
  &Alpaca-7B&6.65&\textbf{92.4 (0.070)}& 7.82& \textbf{75.9 (0.051)}& 6.65&\textbf{90.0 (0.027)}\\
  &Alpaca-13B&6.28&\textbf{89.2 (0.053)}& 6.26& 67.0 (0.043)& 5.59&\textbf{85.4 (0.028)}\\
 &GPT-4o&                                                                   6.11& \textbf{98.2 (0.041)}& 6.12& 95.7 (0.011)& 5.71&\textbf{91.0 (0.026)}\\
\bottomrule
\end{tabular}
}
\caption{Compression ratio with POS (CR-POS) reported for each model-generated output over a random sample (n=500) of the Rotten Tomatoes, Cochrane, and CNN/DM datasets using greedy decoding, and the prompt \texttt{``Write a short summary"}. For Cochrane, we use the prompt \texttt{``Write a meta-analysis"} to match the task. Larger values in CR-POS indicate \textit{less} diversity in the sequences.  We report the percentage of generated outputs with at least 1 template of size $n=6$, and the rate of templates-per-token in parentheses (avg. num. templates per summary normalized by avg. length). Models producing higher templates-per-token than the human-written references are marked in \textbf{bold}.}
\label{tab:cr_pos_greedy_rt}
\end{centering}

\end{table*}

\begin{figure}[t]
    \centering
    \resizebox{0.5\textwidth}{!}{
    \includegraphics{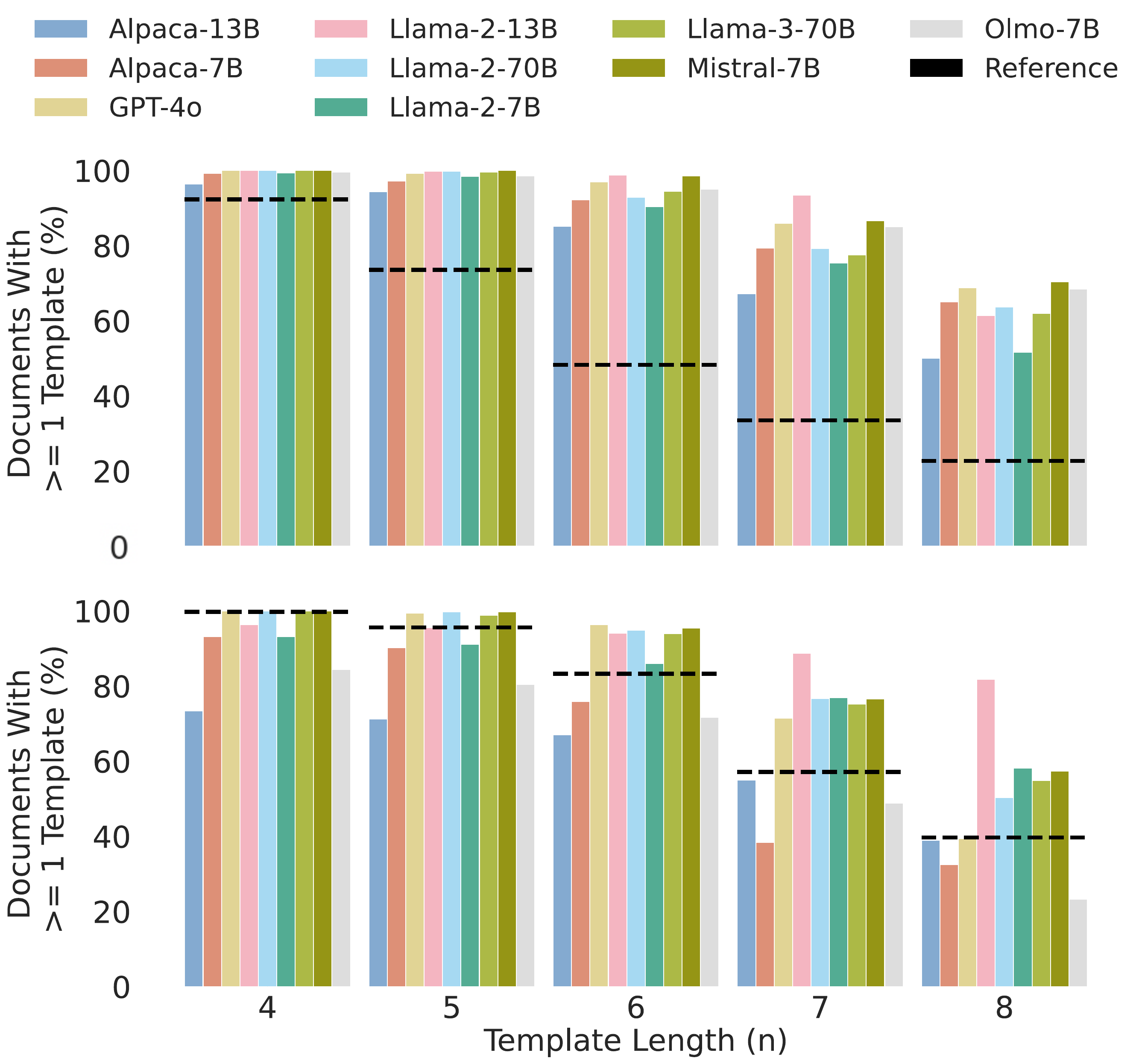}
    }
    \caption{Incidence of generated text with at least 1 template of sizes $n=4, 5, 6, 7, 8$  over the (a) Rotten Tomatoes and (b) Cochrane datasets. Longer templates appear less frequently but at higher rates in model-generated text than in human-written references (dashed lines).}
    \label{fig:templates_length}
\end{figure}
\section{Templates in Closed-Source Models}
\label{sec:templates-in-text-closedsrc}
With OLMo, we find that 75\% of templates are found at high frequencies in the pre-training data (\S\ref{sec:pre-train-data}). 
Most available models however do not release their pre-training data.
Here we evaluate the incidence of templates in other closed-source models, which we define as models that do not release their training data. 
Addressing RQ1, we characterize the rates of templatic texts in these models, and posit that templates may be indicators of the pre-training data sources models are trained on. 

\paragraph{Summarization} 
We report the template rate and templates-per-token that appear in text generated by models in three summarization tasks: movie reviews, biomedical evidence, and news (Table~\ref{tab:cr_pos_greedy_rt}). 

We find that, on average in the Rotten Tomatoes dataset, 95\% of outputs contain templates of length $n=6$ across different model types and sizes. 
This is in contrast to human-written reference and input documents, which contain templates of the same size on average in 38\% of cases. We find a similar trend for templates of length $n = [4, 8]$ (Figure \ref{fig:templates_length}). 

While the average number of templates is higher in model-generated output, this could be attributed to models simply producing lengthier texts than the human written references.
To quantify this, we also compute the template-per-token as a length normalized value capturing the average templates per summary. Even controlling for length, most models produce more templates per token than human authors, as shown in Table \ref{tab:cr_pos_greedy_rt}.

The CNN/DM datasets show similar trends, but with lower rates of templates (average 89.6\% contain templates) compared to the Rotten Tomatoes dataset. 
In contrast, the percentage of templates is high for model-generated (average 88.3\%) and human-written references (83.3\%) in the Cochrane dataset. 
This owes to the nature of meta-analysis texts, which are formulaic \citep{Higgins2010CochraneHF}.\footnote{Table \ref{appx:cochrane_refs} in the appendix provides examples of the human-written references} 

Figure~\ref{fig:templates_length} illustrates the rate of templates for each model as the template length grows from length $n=4$ to $8$ for Rotten Tomatoes and Cochrane. 
For Rotten Tomatoes (and CNN; \Cref{appx:cnn_dm}), all models produce templates at higher rates than human-written summaries across all template lengths. 
With Cochrane, template lengths $\geq 6$ show the majority of models produce higher rates of templates than human authored references. This indicates that differences between templatedness in human-authored references and LLM summaries surface only at longer template lengths. 

The BooookScore dataset provides text generated by models using long-document summarization strategies. We report results over the hierarchical strategy, where the final summary is merged together from smaller summarized chunks. Similar to the other summarization tasks, we observe high rates of templates across all the models available in this dataset (Table~\ref{tab:cr_pos_booookscore}).
\begin{table}[t]
\begin{centering}
\resizebox{0.4\textwidth}{!}{
\begin{tabular}{lrr}
  &  \multicolumn{2}{c}{\textbf{BooookScore, Hierarchical}}\\
\toprule
 \textbf{Model}& \textbf{\begin{tabular}[r]{@{}r@{}} CR: \\POS\end{tabular}}&\textbf{\textbf{\begin{tabular}[r]{@{}r@{}}$\geq1$ Templates\\\% ($n=6$)\end{tabular}}}    \\ \midrule
  
  Claude-2048& 5.63&95.0 (0.010)\\
 Claude-88000& 5.60&94.0 (0.004)\\
 ChatGPT-2048& 6.17&100.0 (0.017)\\
 GPT4-2048& 6.04&100.0 (0.013)\\
 GPT4-4096& 6.01&99.0 (0.013)\\

  Mixtral-2048& 6.01&100.0 (0.017)\\\bottomrule
\end{tabular}
}
\caption{Compression ratio with POS (CR-POS) reported for  the BooookScore dataset. We report the percentage of generated outputs with at least 1 template of size $n=6$, and the rate of templates-per-token in parentheses.}
\label{tab:cr_pos_booookscore}
\end{centering}

\end{table}

\paragraph{Effect of Model Size on Template Rates} 
Table~\ref{tab:cr_pos_greedy_rt} reports differences in the rate of templates between different sizes of Llama-2 and Alpaca.
In the case of Alpaca, the larger model yields outputs with less repetition and fewer templates. 
With the Llama-2 and Llama-3 models, we observe a surprising trend as model size increases: CR-POS and average text length decrease, however the rate of summaries that contain one or more templates stays the same (and increases slightly in some cases). 
These results indicate that larger models do not necessarily produce less templated outputs. 
The templates-per-token value further supports this, showing an increase in template rate (per token) for larger models.

\begin{figure}[t]
    \centering
    \resizebox{0.46\textwidth}{!}{
    \includegraphics{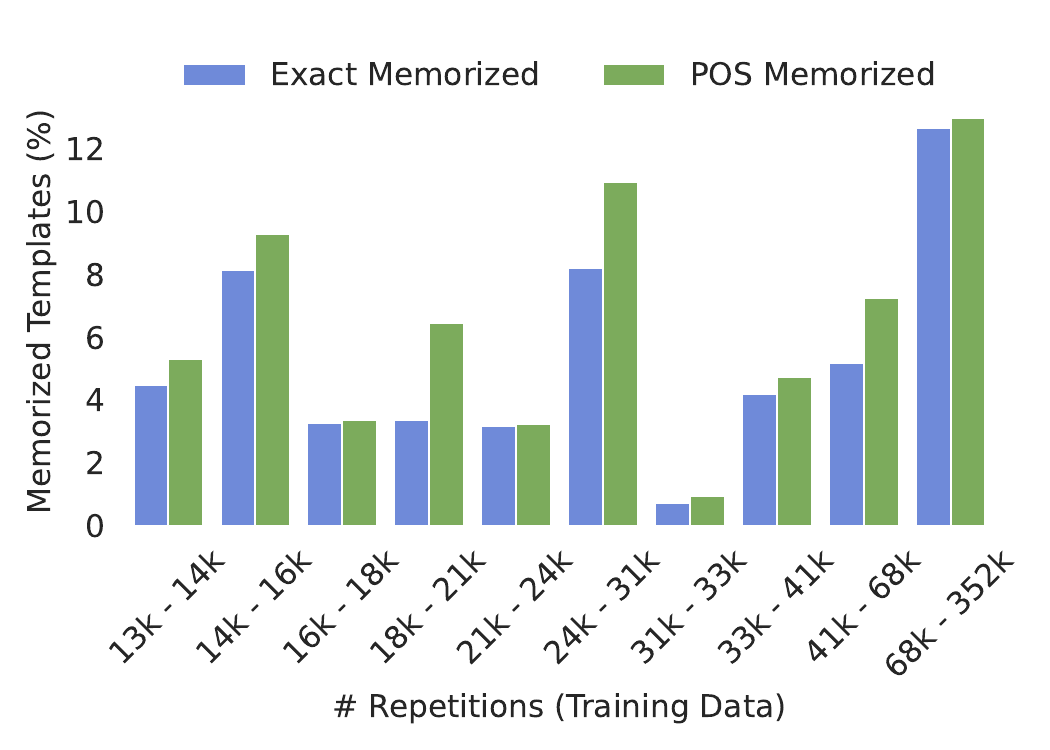}
    }
    \caption{Percent memorized POS sequences, stratified by frequency in training data.}
    \label{fig:memorization}
\end{figure}
\begin{figure}[t]
    \centering
    \resizebox{0.5\textwidth}{!}{
    \includegraphics{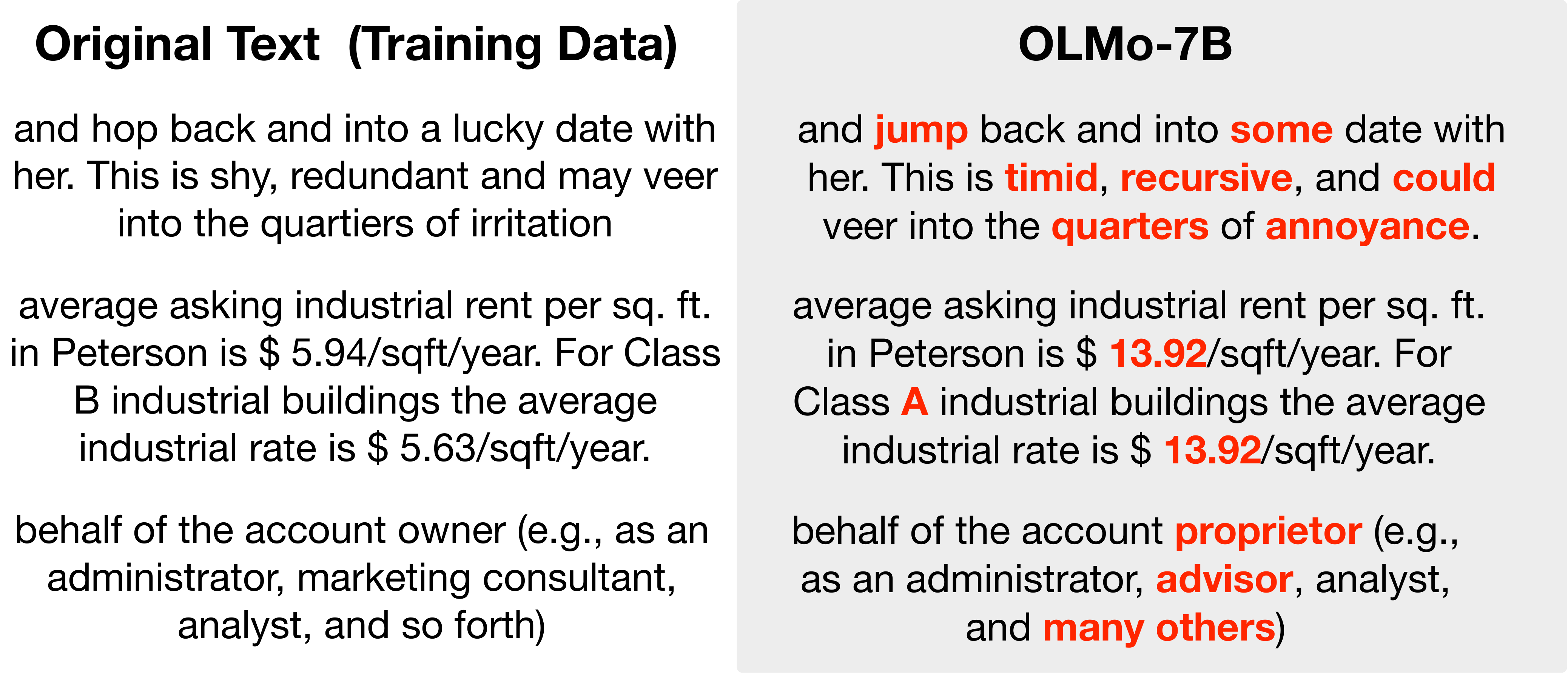}
    }
    \caption{Sampled sentences from OLMo-7B prompted with a prefix of training data. The model substitutes synonyms or numbers.}
    \label{fig:ex_memorization}
\end{figure}
\section{Style Memorization}
\label{sec:memorization}
Past work has shown that models memorize portions of pre-training data. \citet{carlini2022quantifying} show a lower bound of 1\% verbatim memorized data. We next show how our syntactic template analysis can be used to evaluate how much models memorize from pre-training data, beyond strict token sequence matches (\textbf{RQ3}).

\paragraph{Definition: Exact-Text Memorization} We borrow the definition for extractable memorization from \citet{carlini2022quantifying}.
A string $s$ is memorized by a model $\mathcal{M}$ if, when prompted with context $p$, $\mathcal{M}(p)$ produces a string $g$ that is an exact text match to the source string $s$ under greedy decoding. 

\paragraph{Definition: Style (POS) Memorization} For style memorization, we follow the same definition for Exact-Text, but modified to operate over POS (rather than token) sequences to capture instances of syntactic ``style''. 
Specifically, given a POS tagger $f$, sequence $f(s)$ is memorized by $\mathcal{m}$ if, when prompted with context $p$, $\mathcal{m}(p)$ produces a sequence $g$ such that $f(g)$ is an exact match to the source string $f(s)$. 

\subsection{Experimental Setup}
We follow a similar experimental setup as \citet{carlini2022quantifying}, focusing on creating sampled datasets that contain $n$-grams repeated in the pre-training dataset. 
We use \textsc{WIMBD} to build our subset over the Dolma dataset and return the top 50k most common 100-grams in the Dolma dataset from \S~\ref{sec:pre-train-data}.
For each 100-gram, we tokenize the sequence with NLTK and truncate to 50 tokens \citep{carlini2022quantifying}. 
Following the setup for extracting memorized sequences, we prompt OLMo-7B with 50 tokens of the training data sequence and generate a maximum of 1000 tokens using greedy decoding.
We apply NLTK's POS model to tag the original string $s$ and the model-generated string $g$. 

\subsection{Results}
We randomly sample 10k documents and look at the fraction of memorized outputs based on exact-match and the POS sequence (``style'') memorization using the \texttt{diversity} package (e.g., Fig. \ref{fig:ex_method}). We average the fraction memorized over 1,000 seeds for sampling the datasets. 

On average, the POS memorization definition finds 6.4\% ($\pm$ 0.7) memorized, whereas exact text match only reports 5.3\% ($\pm$ 0.6) memorized of the training dataset. \Cref{fig:memorization} shows the percent templates memorized stratified by frequency of the 100-gram in the training dataset. We divide the sampled data point into 10 buckets each containing 4,138 samples with counts that fall in each bucket. In all buckets, POS memorization captures a higher rate of memorized sequences.

The implications of our looser definition of memorization allows us to capture instances of memorization where exact tokens may be substituted during generation, but where an output span is nonetheless structurally the same as a source string. 
Note that this method will by default also capture duplicate text in addition to softly memorized sequence.
In \Cref{fig:ex_memorization}, we provide sampled examples of substitutions that occur that are \textit{not} captured by exact-memorization definitions, yet demonstrate that the particular style of that training point has been memorized.
We find that these cases often include synonym swaps, or different numbers being generated. 

\section{Conclusions} In this work, we introduce syntactic templates as a framework for analyzing subtle repetitive characteristics in model-generated text. We show that this analysis can also extend to human-written references and downstream tasks, and find that the pre-training data contains many of these identified templates. We show that  evaluating repetition in parts-of-speech sequences is useful for detecting subtle types of data ``memorization''. 
Our hope is that this work inspires additional research into characterizing where (in data) observed stylistic patterns in LLM outputs originate.

\section*{Limitations}
There are a few limitations to this work that we address here. 

First, this type of analysis requires an entire corpus that is representative of a text source. For paid models, this can be costly to obtain. For large datasets, this can be resource intensive. These considerations provide a potential barrier based on available resources.

Second, we use third party tools to tag our text abstractions; however these tools are deterministic, but can contain errors in the tags they assign to sequences, particularly if a sequence contains text from another language.
We assume that the majority of the text we analyze is in English, and that any errors are superseded by the frequency of common templates.

Finally, this work only examines English texts, in part due to availability of datasets at the scale necessary to evaluate models. 

\section*{Acknowledgements}
We thank Niloofar Mireshghallah for guidance on the memorization experiments. We also thank Kyle Lo, and Luca Soldaini for their advice and assistance on accessing OLMo and pretraining data.  
This work was supported in part by the National Science Foundation (NSF) grants IIS 2211954, IIS 2145479, IIS 2107524.

\bibliography{anthology,custom}
\bibliographystyle{acl_natbib}
\clearpage

\appendix
\section{Constituency Trees}
\label{appx:consituents}
For constituency parsing, we use the Stanza library \citep{bauer-etal-2023-semgrex}, and linearize sequences using a breadth-first search approach. \Cref{tab:ctree} shows some examples of linearized constituency trees and their matching text. 
\begin{table}[]
\resizebox{0.5\textwidth}{!}{
\begin{tabular}{@{}lll@{}}
\toprule
\textbf{Template (Constituents)}                                                                                                & \textbf{Frequency}    & \textbf{Matched Text}                     \\ \midrule
\multirow[t]{4}{*}{\begin{tabular}[t]{@{}l@{}}(VP (VB ) (NP (NP (DT )\\ (JJ ) (NN )) (PP (IN ) (NP)\end{tabular}}  & \multirow[t]{4}{*}{458}  & Trace the intellectual history of ancient \\
                                                                                                                &                       & Examine the early history of automobiles  \\
                                                                                                                &                       & guarantee a seamless ascent into another  \\ 
                                                                                                                &                       & reach a broad audience of buyers          \\
\multirow[t]{4}{*}{\begin{tabular}[t]{@{}l@{}}(PP (IN ) (NP (NP (DT ) (JJ )\\  (NN )) (PP (IN ) (NP)\end{tabular}} & \multirow[t]{4}{*}{948}  & as a key component of your                \\
                                                                                                                &                       & by a high abundance of free               \\
                                                                                                                &                       & across a global range of cultures         \\
                                                                                                                &                       & for the comprehensive study of the        \\
\multirow[t]{3}{*}{\begin{tabular}[t]{@{}l@{}}(DT ) (JJ ) (NN ))\\  (PP (IN ) (NP)\end{tabular}}                   & \multirow[t]{3}{*}{1680} & a strong grasp of various                 \\
                                                                                                                &                       & a solid understanding of                  \\
                                                                                                                &                       & a radical change in                       \\ \bottomrule
\end{tabular}
}
\label{tab:ctree}
\caption{Example templates using constituency trees over the Cosmopedia dataset.}
\end{table}

\section{CNN/DM Trends}
\label{appx:cnn_dm}
\begin{figure}
    \centering
    \resizebox{0.5\textwidth}{!}{
    \includegraphics{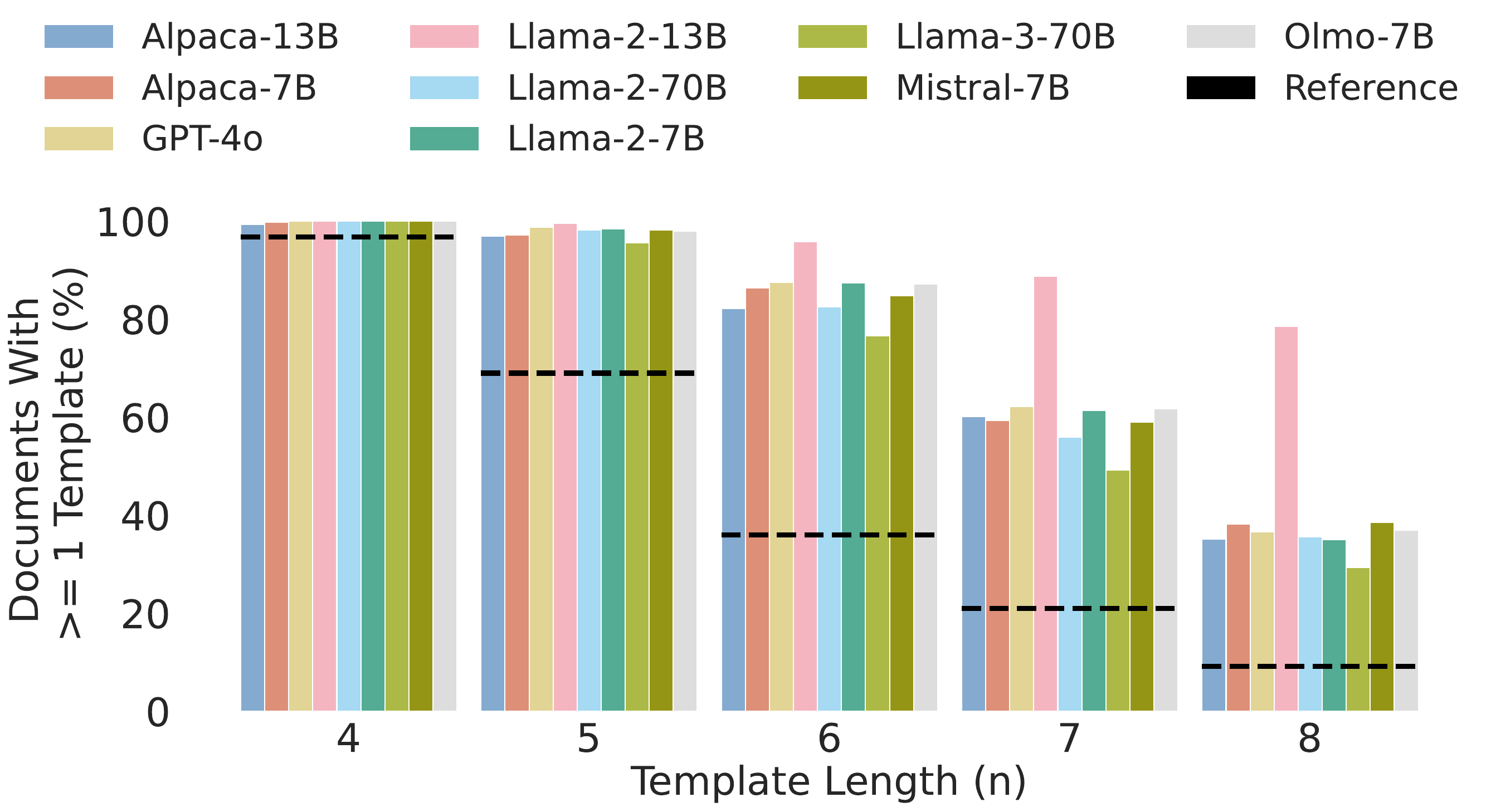}
    }
    \caption{Incidence of generated text with at least 1 template of sizes $n=4, 5, 6, 7, 8$  over the CNN/DM dataset.}
    \label{fig:templates_length_cnn}
\end{figure}

\Cref{fig:templates_length_cnn} demonstrates the same trend of high templatic text in the model generated text compared to human authored references.

\begin{table}
    \centering
    \resizebox{0.5\textwidth}{!}{
    \begin{tabular}{l}
    \toprule
        \textbf{Prompts, CNN/DM}\\ \midrule
        \textbf{1}\ \ \ Please write a summary.                                                                                  \\
        \textbf{2}\ \ \ Please write a summary of the article.\\
        \textbf{3}\ \ \ Summarize.                                                                                               \\
        \textbf{4}\ \ \ Write a short summary.                                                                                   \\
        \textbf{5}\ \ \ Write a short summary and be creative.                       \\
        \textbf{6}\ \ \ Write a meta-analysis.                                                                                   \\
        \textbf{7}\ \ \ Write an aggregate summary based on the above facts.\\ \bottomrule
    \end{tabular}
    }
    \caption{Prompts used for the CNN/DM summarization task.}
    \label{tab:prompts_cnn}
\end{table}

\begin{table}
    \centering
    \resizebox{0.5\textwidth}{!}{
    \begin{tabular}{l}
    \toprule
        \textbf{Prompts, Cochrane}\\ \midrule
        \textbf{1}\ \ \ Please write a summary.                                                                                  \\
        \textbf{2}\ \ \ Please write a summary of the evidence.\\
        \textbf{3}\ \ \ Summarize.                                                                                               \\
        \textbf{4}\ \ \ Write a short summary.                                                                                   \\
        \textbf{5}\ \ \ Write a short summary and be creative.                       \\
        \textbf{6}\ \ \ Write a meta-analysis.                                                                                   \\
        \textbf{7}\ \ \ Write an aggregate analysis based on the above evidence.\\ \bottomrule
    \end{tabular}
    }
    \caption{Prompts used for the Cochrane meta-analysis task.}
    \label{tab:prompts_cochrane}
\end{table}

\section{Cochrane References}
\label{appx:cochrane_refs}
\begin{figure}
    \centering
    \resizebox{0.5\textwidth}{!}{
    \includegraphics{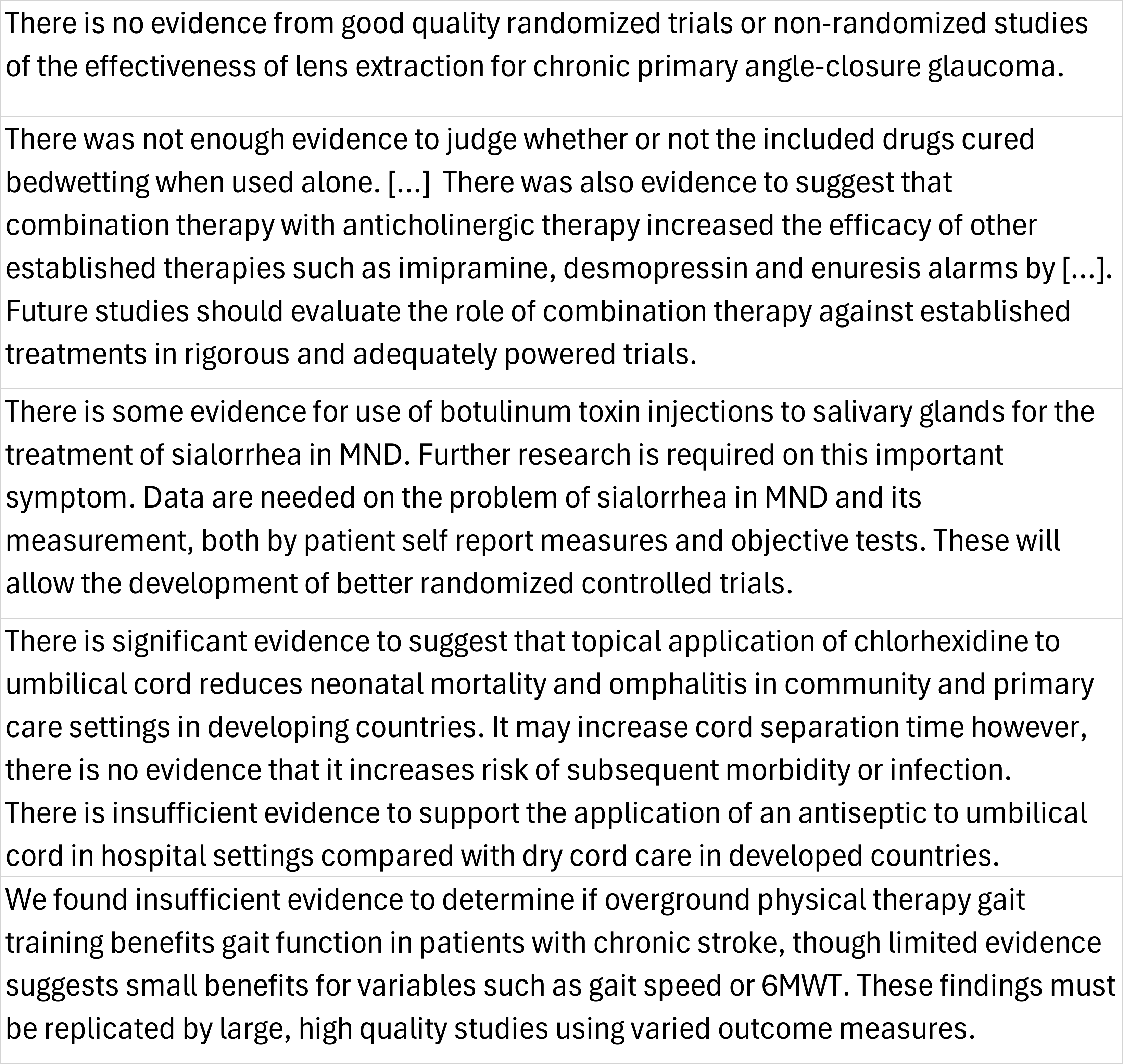}
    }
    \caption{Sample of Cochrane (Human-Authored) references.}
    \label{fig:cochrane}
\end{figure}

Cochrane systematic reviews follow guidelines for how they should be written. \Cref{fig:cochrane} shows an example of human-authored Cochrane text.

\begin{table}
    \centering
    \resizebox{0.5\textwidth}{!}{
    \begin{tabular}{l}
    \toprule
        \textbf{Prompts, Rotten Tomatoes}\\ \midrule
        \textbf{1}\ \ \ Please write a summary.                                                                                  \\
        \textbf{2}\ \ \ Please write a summary of the movie reviews.                \\
        \textbf{3}\ \ \ Summarize.                                                                                               \\
        \textbf{4}\ \ \ Write a short summary.                                                                                   \\
        \textbf{5}\ \ \ Write a short summary and be creative.                       \\
        \textbf{6}\ \ \ Write a meta-analysis.                                                                                   \\
        \textbf{7}\ \ \ Write an aggregate movie review based on the above reviews. \\ \bottomrule
    \end{tabular}
    }
    \caption{Prompts used for the Rotten Tomatoes summarization task.}
    \label{tab:prompts}
\end{table}

\section{Choice of Prompt}
\label{appx:prompts}
The choice of prompt can impact the incidence of templates. Here, we provide an analysis of 7 different prompts for the downstream summarization task. 
We find that shorter and more generic prompts result in lower performance. 
It is possible that with the shorter and more generic instructions instructions, there may be a higher overlap in template types across models.

\begin{table}[t]
\begin{centering}
\resizebox{0.5\textwidth}{!}{
\begin{tabular}{@{}llllllll|c}
\toprule
               \textbf{Model}& \multicolumn{7}{c}{\textbf{$\geq1$ Templates \% ($n=6$)}}&\\ \midrule
 \textbf{Prompt}& \textbf{1}& \textbf{2}& \textbf{3}&\textbf{4}& \textbf{5}& \textbf{6}&\textbf{7} &$\Delta$\\ \midrule
OLMo-7B        &                   98.1&                   94.4&                   92.5& 97.0 &                   96.2&                   97.2&                   97.2 &5.6\\
Mistral-7B     &                   98.5&                   98.5&                   97& 99.6 &                   99.5&                   100&                   100 &3.0\\
Llama-2-7B     &                   91.5&                   94.5&                   88.9& 93.0&                   94.8&                   95.4&                   94.9 &6.5\\
Llama-2-13B    &                   99.1&                   99.8&                   94.9& 99.0&                   98.0&                   98.0&                   98.6 &4.9\\
Llama-2-70B    &                   96.2&                   96.8&                   97.2& 99.3&                   98.8&                   99.6&                   99.2 &3.4\\
Llama-3-70B    &                   95.7&                   97.3&                   93.7& 99.2&                   97.8&                   99.8&                   99.4 &6.1\\
Alpaca-7B      &                   92.3&                   87.5&                   86.8& 92.4&                   92.7&                   84.1&                   91.6 &8.6\\
Alpaca-13B     &                   92.5&                   91.0&                   90.3& 89.2&                   91.6&                   94.5&                   94.4 &5.3\\
GPT-4o&                   98.2&                   98.0&                   95.2&                   97.0&                   99.6&                   100.0&                    99.8&4.8\\ \bottomrule
\end{tabular}
}
\caption{
Percentage of generated summaries with at least 1 template of length $n=6$ under different instruction prompts. Some instruction prompts result is more templates outputs for certain models.}
\label{tab:cr_pos_greedy_rt_all_prompts}
\end{centering}

\end{table}

The choice of prompt can affect the rate of templates. Appendix \Cref{tab:cr_pos_greedy_rt_all_prompts} shows the rate of templates with different prompts while the choice of prompt impacts the rate of templates, the incidence remains on average higher than 90\%.

\section{Template Measures, Text Length}
For completeness, we provide the full tables for all datasets that include the average text length. In the case of Booookscore, we also provide results over the incremental dataset.
\label{appx:text_length}
\begin{table*}[t]
\begin{centering}
\resizebox{0.9\textwidth}{!}{
\begin{tabular}{l@{}llll|lll|lll}
 & & \multicolumn{3}{c}{\textbf{\begin{tabular}[c]{@{}l@{}} Rotten Tomatoes\end{tabular}}}& \multicolumn{3}{c}{\textbf{Cochrane}}& \multicolumn{3}{c}{\textbf{CNN/DM}}\\
\toprule
 &\textbf{Model}& \textbf{\begin{tabular}[c]{@{}l@{}} CR: \\POS\end{tabular}}&\textbf{\begin{tabular}[c]{@{}l@{}} Avg. Text \\Length\end{tabular}}&\textbf{\textbf{\begin{tabular}[c]{@{}l@{}}$\geq1$ Templates\\\% ($n=6$)\end{tabular}}}   & \textbf{\begin{tabular}[c]{@{}l@{}} CR: \\POS\end{tabular}}& \textbf{\begin{tabular}[c]{@{}l@{}} Avg. Text \\Length\end{tabular}}& \textbf{\textbf{\begin{tabular}[c]{@{}l@{}}$\geq1$ Templates\\\% ($n=6$)\end{tabular}}}   & \textbf{\begin{tabular}[c]{@{}l@{}} CR: \\POS\end{tabular}}& \textbf{\begin{tabular}[c]{@{}l@{}} Avg. Text \\Length\end{tabular}}&\textbf{\textbf{\begin{tabular}[c]{@{}l@{}}$\geq1$ Templates\\\% ($n=6$)\end{tabular}}}   \\\midrule
  &
  Reference& 5.31&25.1&46.4 (0.040) & 5.63& 73.8& 83.3 (0.049)& 5.33& 57.7&36.0 (0.013)\\
  &Input Documents& 5.82&668.2&29.3 (0.001) & 5.96& 1555.3& 98.5 (0.021)& 5.54& 514.7&98.4 (0.020)\\ \midrule
  &
  OLMo-7B     & 6.45&194.4&97.0 (0.041) & 6.53& 158.1& 74.0 (0.030)& 5.83& 177.1&\textbf{91.2 (0.025)}\\
 &Mistral-7B  & 6.29&185.6&\textbf{99.6 (0.043)}& 6.10& 177.2& 99.5 (0.043)& 5.70& 153.0&\textbf{89.9 (0.029)}\\
 &Llama-2-7B  & 6.87&126.5&\textbf{93.0 (0.047)}& 6.43& 151.0& 88.4 (0.042)& 5.71& 153.9&\textbf{90.4 (0.028)}\\
 &Llama-2-13B &                                                                   6.70&117.0&\textbf{99.0 (0.060)}& 6.65& 157.7& \textbf{95.1 (0.052)}& 5.91& 143.5&\textbf{97.4 (0.042)}\\
 &Llama-2-70B &                                                                   6.36&114.2&\textbf{99.3 (0.123)}& 6.51& 324.7& 99.7 (0.042)& 5.69& 138.3&\textbf{87.4 (0.027)}\\
 &Llama-3-70B&     6.39&      106.1&          \textbf{99.2 (0.151)}& 6.50& 387.5& 99.5 (0.030)& 5.66& 132.7&\textbf{83.2 (0.024)}\\
  &Alpaca-7B&6.65&99.4&\textbf{92.4 (0.070)}& 7.82& 98.1& \textbf{75.9 (0.051)}& 6.65& 145.2&\textbf{90.0 (0.027)}\\
  &Alpaca-13B&6.28&93.0&\textbf{89.2 (0.053)}& 6.26& 69.7& 67.0 (0.043)& 5.59& 138.1&\textbf{85.4 (0.028)}\\
 &GPT-4o&                                                                   6.11&203.5& \textbf{98.2 (0.041)}& 6.12& 560.7& 95.7 (0.011)& 5.71& 167.6&\textbf{91.0 (0.026)}\\
\bottomrule
\end{tabular}
}
\caption{Compression ratio with POS (CR-POS) reported for each model-generated output over a random sample (n=500) of the Rotten Tomatoes, Cochrane, and CNN/DM datasets using greedy decoding, and the prompt \texttt{``Write a short summary"}. For Cochrane, we use the prompt \texttt{``Write a meta-analysis"} to match the task. Larger values in CR-POS indicate \textit{less} diversity in the sequences.  We report the percentage of generated outputs with at least 1 template of size $n=6$, and the rate of templates-per-token in parentheses (avg. num. templates per summary normalized by avg. length). Models producing higher templates-per-token than the human-written references are marked in bold.}
\label{tab:cr_pos_greedy_rt_len}
\end{centering}

\end{table*}

\begin{table*}
\begin{centering}
\resizebox{0.75\textwidth}{!}{

\begin{tabular}{llll|lll}
  &  \multicolumn{3}{c}{\textbf{Open Generation}}&  \multicolumn{3}{c}{\textbf{Rotten Tomatoes}}\\
\toprule
 \textbf{Decoding Strategy}& \textbf{\begin{tabular}[c]{@{}l@{}} CR: \\POS\end{tabular}}& \textbf{\begin{tabular}[c]{@{}l@{}} Avg. Text \\Length\end{tabular}}&\textbf{\textbf{\begin{tabular}[c]{@{}l@{}}$\geq1$ Templates\\\% ($n=6$)\end{tabular}}}    &  \textbf{\begin{tabular}[c]{@{}l@{}} CR: \\POS\end{tabular}}& \textbf{\begin{tabular}[c]{@{}l@{}} Avg. Text \\Length\end{tabular}}& \textbf{\textbf{\begin{tabular}[c]{@{}l@{}}$\geq1$ Templates\\\% ($n=6$)\end{tabular}}}    \\\midrule
 Greedy& 702.8& 48.3& 100.0 (0.065)& 6.45& 194.4&97.0 (0.041) \\ \midrule
 Default Sampling& 5.81& 479.2&75.5 (0.009)&  6.33& 194.2& 96.6 (0.041)\\
\texttt{temp 0.8} & 6.74& 517.9&71.8 (0.007)&  6.26& 191.1& 96.6 (0.043)\\
\texttt{temp 0.85}& 6.48& 506.8& 74.0 (0.007)& 6.22& 188.5&96.6 (0.041)\\
\texttt{temp 0.9}& 6.22& 497.3& 73.4 (0.009)& 6.17& 190.3&97.4 (0.039)\\
\texttt{temp 0.95}& 5.98& 500.2& 75.4 (0.01)& 6.14& 186.6&97.2 (0.040)\\\midrule
\texttt{top\_p 0.8}& 7.03& 541.0& 76.5 (0.007)& 6.31& 190.2&97.8 (0.041)\\
\texttt{top\_p 0.85}& 6.71& 526.3&71.0 (0.007)&  6.27& 190.3& 96.6 (0.041)\\
\texttt{top\_p 0.9}& 6.50& 495.1&75.3 (0.008)&  6.22& 190.9& 96.2 (0.039)\\
\texttt{top\_p 0.95}& 6.17& 513.0&77.2 (0.009)&  6.31& 192.1& 95.8 (0.041)\\\bottomrule
\end{tabular}
}
\caption{Compression ratio with POS (CR-POS), average text length and percentage of generated outputs with at least 1 template of size $n=6$, when varying \texttt{temperature} and \texttt{top\_p} for OLMo-7B decoding. }
\label{tab:cr_pos_sample_length}
\end{centering}

\end{table*}
\begin{table*}
\begin{centering}
\resizebox{0.8\textwidth}{!}{
\begin{tabular}{lrrr|rrr}
  &  \multicolumn{3}{c}{\textbf{BooookScore, Hierarchical}}&  \multicolumn{3}{c}{\textbf{BooookScore, Incremental}}\\
\toprule
 \textbf{Model}& \textbf{\begin{tabular}[r]{@{}r@{}} CR: \\POS\end{tabular}}& \textbf{\begin{tabular}[r]{@{}r@{}} Avg. Text \\Length\end{tabular}}&\textbf{\textbf{\begin{tabular}[r]{@{}r@{}}$\geq1$ Templates\\\% ($n=6$)\end{tabular}}}    &  \textbf{\begin{tabular}[r]{@{}r@{}} CR: \\POS\end{tabular}}& \textbf{\begin{tabular}[r]{@{}r@{}} Avg. Text \\Length\end{tabular}}& \textbf{\textbf{\begin{tabular}[r]{@{}r@{}}$\geq1$ Templates\\\% ($n=6$)\end{tabular}}}    \\ \midrule
  
  Claude-2048& 5.63& 461.9&95.0 (0.010)&  5.71& 575.4& 86.0 (0.007)\\
 Claude-88000& 5.60& 487.9&94.0 (0.004)&  5.60& 439.5& 96.0 (0.010)\\
 ChatGPT-2048& 6.17& 613.8&100.0 (0.017)&  5.76& 439.2& 95.0 (0.011)\\
 GPT4-2048& 6.04& 706.5&100.0 (0.013)&  5.95& 721.0& 99.0 (0.009)\\
 GPT4-4096& 6.01& 855.4&99.0 (0.013)&  6.05& 1,013.2& 100.0 (0.010)\\

  Mixtral-2048& 6.01& 619.2&100.0 (0.017 )&  5.59& 496.1& 99.0 (0.012)\\\bottomrule
\end{tabular}
}
\caption{Compression ratio with POS (CR-POS) reported for  the BooookScore dataset. We report the percentage of generated outputs with at least 1 template of size $n=6$, and the rate of templates-per-token in parentheses.}
\label{tab:cr_pos_booookscore_length}
\end{centering}

\end{table*}

\begin{table*}
\begin{centering}
\resizebox{0.45\textwidth}{!}{
\begin{tabular}{llll}
\toprule
 \textbf{Dataset}& \textbf{\begin{tabular}[c]{@{}l@{}} CR: \\POS\end{tabular}}&\textbf{\begin{tabular}[c]{@{}l@{}} Avg. Text \\Length\end{tabular}}&\textbf{\textbf{\begin{tabular}[c]{@{}l@{}}$\geq1$ Templates\\\% ($n=6$)\end{tabular}}}   \\\midrule
  
 Dolma-100& 5.65& 483.2&82.6 (0.012)\\ 
  Cosmopedia& 5.76 & 768.0&99.1 (0.014)\\\bottomrule
\end{tabular}
}
\caption{CR-POS, template-per-token, and template counts for templates of size $n=6$ reported for OLMo-7B text generated with Cosmopedia Instructions, and 100 sampled tokens from the Dolma dataset, with greedy decoding.}
\label{tab:cr_pos_cosmopedia_length}
\end{centering}

\end{table*}

\end{document}